\documentclass[runningheads]{llncs}
\usepackage{graphicx}
\usepackage{comment}
\usepackage{amsmath,amssymb} 
\usepackage{color}
\usepackage{multirow}
\usepackage{tabularx}
\usepackage{textcomp}
\def\permille{\ensuremath{{}^\text{o}\mkern-5mu/\mkern-3mu_\text{ooo}}}
\newcolumntype{L}[1]{>{\raggedright\arraybackslash}p{#1}}
\newcolumntype{C}[1]{>{\centering\arraybackslash}p{#1}}
\newcolumntype{R}[1]{>{\raggedleft\arraybackslash}p{#1}}
\usepackage{dsfont}
\usepackage{pifont}
\usepackage{hyperref}
\makeatletter
\newcommand{\printfnsymbol}[1]{%
  \textsuperscript{\@fnsymbol{#1}}%
}
\makeatother
\newcommand{\xmark}{\ding{55}}%

\newcommand{\TODO}[1]{\textcolor{red}{(TODO: #1)}}
\newcommand{\xiao}[1]{\textcolor[rgb]{0,0,1}{(Xiao: #1)}}
\newcommand{\fawe}[1]{\textcolor[rgb]{0,0,1}{(Fangyun: #1)}}

\usepackage{amsmath}

\begin{document}
\pagestyle{headings}
\mainmatter
\def\ECCVSubNumber{1162}  

\title{Point-Set Anchors for Object Detection, Instance Segmentation and Pose Estimation}

\titlerunning{Point-Set Anchors}
%
\author{Fangyun Wei\inst{1}\thanks{Equal contribution.} \and
Xiao Sun\inst{1}\printfnsymbol{1} \and
Hongyang Li\inst{2} \and Jingdong Wang\inst{1} \and Stephen Lin\inst{1}}
\authorrunning{F. Wei, X. Sun, H. Li, J. Wang and S. Lin}
%
\institute{
Microsoft Research Asia \\
\email{\{fawe, xias, jingdw, stevelin\}@microsoft.com}\\
\and
Peking University\\
\email{lhy\_ustb@pku.edu.cn}}
\maketitle

\begin{abstract}
A recent approach for object detection and human pose estimation is to regress bounding boxes or human keypoints from a central point on the object or person. While this center-point regression is simple and efficient, we argue that the image features extracted at a central point contain limited information for predicting distant keypoints or bounding box boundaries, due to object deformation and scale/orientation variation. To facilitate inference, we propose to instead perform regression from a set of points placed at more advantageous positions. This point set is arranged to reflect a good initialization for the given task, such as modes in the training data for pose estimation, which lie closer to the ground truth than the central point and provide more informative features for regression. As the utility of a point set depends on how well its scale, aspect ratio and rotation matches the target, we adopt the anchor box technique of sampling these transformations to generate additional point-set candidates. We apply this proposed framework, called Point-Set Anchors, to object detection, instance segmentation, and human pose estimation. Our results show that this general-purpose approach can achieve performance competitive with state-of-the-art methods for each of these tasks. Code is available at \url{https://github.com/FangyunWei/PointSetAnchor}.
\keywords{Object detection, instance segmentation, human pose estimation, anchor box, point-based representation}
\end{abstract}

\section{Introduction}

A basic yet effective approach for object localization is to estimate keypoints. This has been performed for object detection by detecting points that can define a bounding box, e.g., corner points~\cite{law2018cornernet}, and then grouping them together. An even simpler version of this approach that does not require grouping is to extract the center point of an object and regress the bounding box size from it. This method, called CenterNet~\cite{zhou2019objects}, can be easily applied to human pose estimation as well, by regressing the offsets of keypoints instead.

While CenterNet is highly practical and potentially has broad application, its regression of keypoints from features at the center point can be considered an important drawback. Since keypoints might not lie in proximity of the center point, the features extracted at the center may provide little information for inferring their positions. This problem is exacerbated by the geometric variations an object can exhibit, including scale, orientation, and deformations, which make keypoint prediction even more challenging.

\begin{figure*}[t]
\begin{center}
\includegraphics[width=0.6\linewidth]{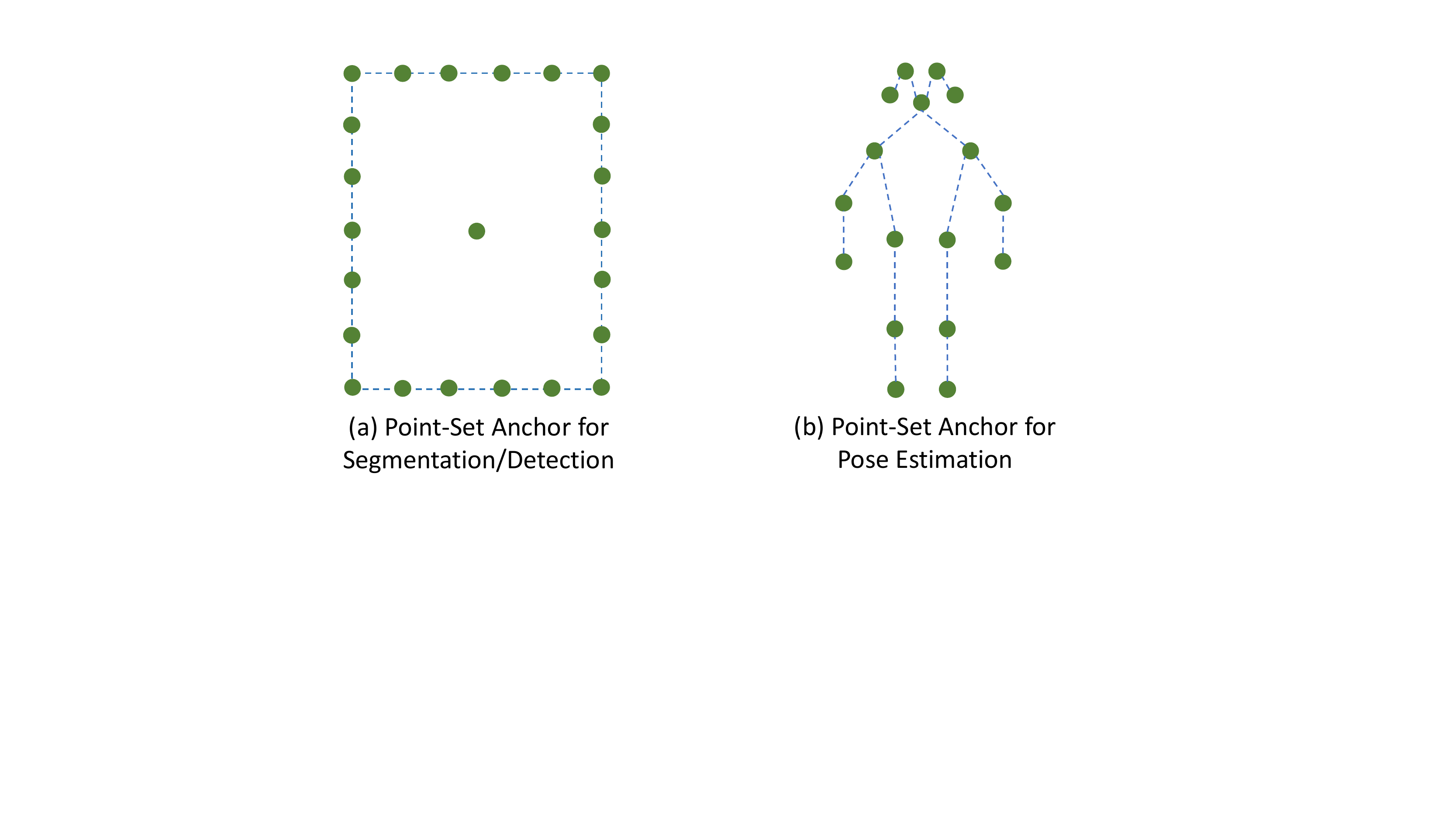}
\end{center}
\vspace{-2ex}
\caption{Illustration of our point-set anchors for instance segmentation, object detection and pose estimation. Instance mask point-set anchors contain an implicit bounding box, and $n$ anchor points are uniformly sampled from the corresponding bounding box. Pose point-set anchors are initialized as the most frequent poses in the training set.}
\label{fig:initial}
\vspace{-0.5cm}
\end{figure*}

In this paper, we propose to address this issue by acquiring more informative features for keypoint regression. Rather than extract them at the center point, our approach is to obtain features at a set of points that are likely to lie closer to the regression targets. This point set is determined according to task. For instance segmentation, the points are placed along the edges of an implicit bounding box. For pose estimation, the arrangement of points follows modes in the pose distribution of the training data, such as that in Fig.~\ref{fig:initial} (b). As a good task-specific initialization, the point set can yield features that better facilitate keypoint localization.

It can be noted that a point set best serves its purpose when it is aligned in scale and aspect ratio with the target. To accomplish this, we adapt the anchor box scheme commonly used in object detection by expressing point sets as {\em point-set anchors}. Like their anchor box counterparts, point-set anchors are sampled at multiple scales, aspect ratios, and image positions. In addition, different point-set configurations may be enumerated, such as different modes in a pose estimation training set. With the generated point-set candidates, keypoint regression is conducted to find solutions for the given task. 

The main contributions of this work can be summarized as:
\begin{itemize}
\item A new object representation named \emph{Point-Set Anchors}, which can be seen as a generalization and extension of classical box anchors. Point-set anchors can further provide informative features and better task-specific initializations for shape regression.
\item A network based on point-set anchors called PointSetNet, which is a modification of RetinaNet~\cite{lin2017focal} that simply replaces the anchor boxes with the proposed point-set anchors and also attaches a parallel regression branch. Variants of this network are applied to object detection, human pose estimation, and also instance segmentation, for which the problem of defining specific regression targets is addressed.
\end{itemize}
It is shown that the proposed general-purpose approach achieves performance competitive with state-of-the-art methods on object detection, instance segmentation and pose estimation.

\section{Related Work}

\noindent\textbf{Object representations.}
In object detection, rectangular anchors~\cite{renNIPS15fasterrcnn,lin2017focal,lin2017feature} are the most common representation used in locating objects. These anchors serve as initial bounding boxes, and an encoding is learned to refine the object localization or to provide intermediate proposals for top-down solutions~\cite{he2017mask,dai2016instance}. 
However, the anchor box is a coarse representation that is insufficient for finer degrees of localization required in tasks such as instance segmentation and pose estimation. 
An alternative is to represent objects in terms of specific points, including center points~\cite{tian2019fcos,zhou2019objects}, corner points~\cite{law2018cornernet,duan2019centernet}, extreme points~\cite{zhou2019bottom}, octagon points~\cite{peng2020deep}, point sets~\cite{yang2019reppoints,yang2019dense}, and radial points~\cite{xie2019polarmask}. These point representations are designed to solve one or two tasks among object detection, instance segmentation and pose estimation. Polygon point based methods, such as corner points~\cite{law2018cornernet,duan2019centernet} and extreme points~\cite{zhou2019bottom}, are hard to apply to instance segmentation and pose estimation due to their restricted shape. While center point representations~\cite{tian2019fcos,zhou2019objects} are more flexible, as offsets from the object center to the corresponding bounding box corners or human joints can be directly predicted, we argue that features extracted from center locations are not as informative as from our task-specific point sets, illustrated in Fig.~\ref{fig:initial}. In addition, how to define regression targets for instance segmentation is unclear for these representations. Another way to perform instance segmentation from a center point is by regressing mask boundary points in radial directions~\cite{xie2019polarmask}; however, radial regressions at equal angular intervals are unsuitable for pose estimation. Other representations such as octagon points or point sets~\cite{yang2019reppoints,yang2019dense} are specifically designed for one or two recognition tasks. The proposed point-set anchors combine benefits from anchor boxes and point representations, and its flexibility makes them applicable to object detection, instance segmentation and pose estimation.

\noindent\textbf{Instance segmentation.}
Two-stage methods~\cite{he2017mask,li2017fully,huang2019mask} formulate instance segmentation in a `Detect and Segment' paradigm, which detects bounding boxes and then performs instance segmentation inside the boxes. Recently, there is much research focused on single-stage instance segmentation since two-stage methods are often slow in practice. PolarMask~\cite{xie2019polarmask} uses a polar representation and $n$ rays at equal angular intervals are emitted from the polar center for dense distance regression. YOLACT~\cite{bolya2019yolact} generates a set of prototype masks and predicts per-instance mask coefficients for linearly combining the prototype masks. ExtremeNet~\cite{zhou2019bottom} detects four extreme points and a center point using a standard keypoint estimation network, which has the disadvantage of a long training time, and then grouping is applied to generate coarse octagonal masks. The recent Deep Snake~\cite{peng2020deep} proposes a two-stage pipeline based on initial contours and contour deformation. Our method is different from Deep Snake in three ways. First, our method for instance segmentation operates in a single stage, without needing proposals generated by detectors. Second, Point-Set Anchors perform mask shape regression directly, in contrast to the iterative deformation in Deep Snake. Finally, our method is evaluated on the challenging MS COCO dataset~\cite{lin2014microsoft} and is compared to state-of-the-art methods in object detection, instance segmentation and pose estimation.

\noindent\textbf{Pose estimation.}
In pose estimation algorithms, most previous works follow the paradigm of estimating a heat map for each joint~\cite{bulat2016human,cao2017realtime,chen2017adversarial,chou2017self,chu2017multi,gkioxari2016chained,he2017mask,pishchulin2016deepcut,insafutdinov2016deepercut,newell2016stacked,xiao2018simple}. The heat map represents the confidence of a joint existing at each position in the image. Despite its good performance, a heat map representation has a few drawbacks such as no end-to-end training, a need for high resolution~\cite{sun2018integral}, and separate steps for joint localization and association. The joint association problem is typically addressed by an early stage of human bounding box detection~\cite{xiao2018simple,he2017mask} or a post-processing step that groups the detected joints together with additionally learned joint relations~\cite{cao2017realtime,newell2017associative}. Recently, a different approach has been explored of directly regressing the offsets of joints from a center or root point~\cite{zhou2019objects}. In contrast to the heat map based approaches, the joint association problem is naturally addressed by conducting a holistic regression of all joints from a single point. However, the holistic shape regression is generally more difficult than part based heat map learning due to optimization on a large, high-dimensional vector space. Nie et al.~\cite{nie2019single} address this problem by factorizing the long-range displacement with respect to the center position into accumulative shorter ones based on the articulated kinematics of human poses. They argue that modeling short-range displacements can alleviate the learning difficulty of mapping from an image representation to the vector domain. We follow the regression based paradigm and propose to address the long-range displacement problem by regressing from a set of points placed at more advantageous positions.

\section{Our Method}

In this section, we first formulate the proposed task-specific point-set anchors. Next, we show how to make use of these point-set anchors for regression-based instance segmentation and pose estimation. Finally, we introduce our PointSetNet, which is an extension of RetinaNet with a parallel branch attached for keypoint regression.

\subsection{Point-Set Anchors}
The point-set anchor $\mathbf{T}$ 
contains an ordered set of points 
that are defined according to the specific task. 
We describe the definitions for the tasks of human pose estimation
and instance segmentation that we will handle.

\vspace{1mm}
\noindent\textbf{Pose point-set anchor.}
We naturally use the human keypoints 
to form the pose point-set anchor.
For example, in the COCO~\cite{lin2014microsoft} dataset,
there are $17$ keypoints,
and the pose point-set anchor is represented
by a $34$-dimensional vector.
At each image position,
we use several point-set anchors.
We initialize the point-set anchors
as the most frequent poses in the training set. 
We use a standard k-means clustering algorithm~\cite{lloyd1982least} to partition all the training poses into $k$ clusters, 
and the mean pose of each cluster is used to form a point-set anchor. Fig.~\ref{fig:initial}(b) illustrates one of the point-set anchors for pose estimation. 

\vspace{1mm}
\noindent\textbf{Instance mask point-set anchor.}
This anchor has two parts: one center point and $n$ ordered anchor points which are uniformly sampled from an implicit bounding box as illustrated in Fig.~\ref{fig:initial}(a). $n$ is a hyper-parameter to control sampling density. Corner points in a mask point-set anchor can also serve as a reference for object detection. At each image position, we form $9$ point-set anchors by varying the scale and aspect ratios of the implicit bounding box.

\subsection{Shape Regression}
\label{sec:shape_reg}
In this work, instance segmentation, object detection and pose estimation 
are treated as a shape regression problem. 
We represent the object by a shape $\mathbf{S}$,
i.e., a set of $n_s$ ordered points
$\textbf{S}=\{S_{i}\}_{i=1}^{n_s}$, 
where $S_{i}$ represents the $i$-th \emph{polygon vertex} for an instance mask, $i$-th \emph{corner vertex} for a bounding box, or the $i$-th \emph{keypoint} for pose estimation. 
Instead of regressing the shape point locations from the object center, we employ $\mathbf{T}$ as a reference for shape regression.
Our goal is to regress the offsets $\Delta\mathbf{T}$
from the point-set anchor $\mathbf{T}$ to the shape $\mathbf{S}$.

\vspace{1mm}
\noindent\textbf{Offsets for pose estimation.}
Each keypoint in the human pose estimation task represents a joint with semantic meaning, e.g., head, right elbow or left knee. 
We use $17$ keypoints as the shape $\mathbf{S}$ 
and the offsets are simply the difference:
$\Delta\mathbf{T} = \mathbf{S} - \mathbf{T}$.

\begin{figure*}[t]
\begin{center}
\includegraphics[width=0.95\linewidth]{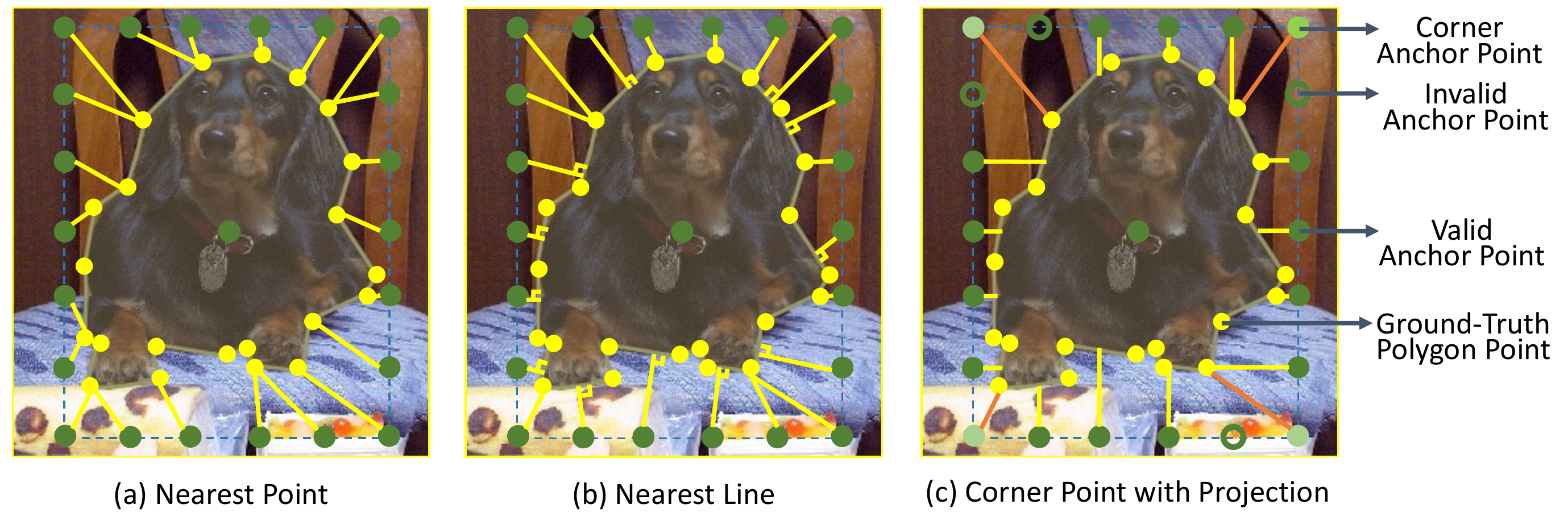}
\end{center}
\vspace{-2ex}
\caption{Illustration of three matching strategies between a point-set anchor and a ground-truth mask contour for instance segmentation. Yellow solid dots represent polygon points of the ground-truth mask. Green solid dots and green hollow dots denote valid and invalid anchor points, respectively. Orange and yellow lines indicate correspondences for corner and non-corner anchor points, respectively. Only valid anchor points are considered for training and inference.}
\label{fig:match_methods}
\vspace{-0.5cm}
\end{figure*}

\vspace{1mm}
\noindent\textbf{Offsets for instance segmentation.}
The shape $\mathbf{S}$ of an instance mask also contains 
a set of ordered points,
and the number of points might be different 
for different object instances.
The point-set anchor $\mathbf{T}$ is defined for all instances
and contains a fixed number of points.
To compute the offsets $\Delta\mathbf{T}$,
we find the matching points $\mathbf{T}^*$, 
each of which has a one-to-one correspondence to each point in $\mathbf{T}$, from the shape $\mathbf{S}$,
and then $\Delta \mathbf{T} = \mathbf{T}^* - \mathbf{T}$.
The matching strategies, illustrated in Fig.~\ref{fig:match_methods},
are described as follows.

\begin{itemize}
    \item \textbf{Nearest point.} The matching target of each point in the point-set anchor
    $\textbf{T}$ is defined as the nearest polygon point in $\textbf{S}$ based on $L_{1}$ distance. Thus, a single polygon point may be assigned to several anchor points, one anchor point, or none of them.
    \item \textbf{Nearest line.} We treat the mask contour \textbf{S} as a sequence of $n_s$ line segments instead of $n_s$ discrete polygon vertices. Each of the anchor points is projected to all the polygon edges, and the closest projection point is assigned to the corresponding anchor point.
    \item \textbf{Corner point with projection.} We first find the targets of the four corner anchor points by the Nearest Point strategy, which are then used to subdivide the mask contour into four parts, i.e., top, right, bottom and left parts, that are to be matched with anchor points on the corresponding side. For each of the four parts, the target of each anchor point is the nearest intersection point between the horizontal (for left and right parts) or vertical (for top and bottom parts) projection line and the line segments of the mask contour. If a matched point lies outside of the corresponding contour segment delimited by the matched corner points, we mark it as invalid and it is ignored in training and testing. The remaining anchor points and their matches are marked as valid for mask regression learning.
\end{itemize}

\noindent\textbf{Offsets for object detection.} The bounding box shape \textbf{S} in object detection can be denoted as two keypoints, i.e., the top-left and bottom-right point. The offsets $\Delta\textbf{T}$ are the distance from the target points to the corresponding corner points in mask point-set anchors.

\vspace{1mm}
\noindent\textbf{Positive and negative samples.} In assigning positive or negative labels to point-set anchors, we directly employ IoU for object detection and instance segmentation, and Object Keypoint Similarity (OKS) for pose estimation. Formally, in instance segmentation and object detection, a point-set anchor is assigned a positive label if it has the highest IoU for a given ground-truth box or an IoU over 0.6 with any ground-truth box, and a negative label if it has an IoU lower than 0.4 for all ground-truth boxes. In practice, we use the IoU between the implicit bounding box of the mask point-set anchors and ground-truth bounding boxes instead of masks in instance segmentation, for reduced computation. For human pose estimation, a point-set anchor is assigned a positive label if it has the highest OKS for a given ground-truth pose or an OKS over 0.5 with any ground-truth pose, and a negative label if it has OKS lower than 0.4 for all ground-truth poses.

\vspace{1mm}
\noindent\textbf{Mask construction.} In instance segmentation, a mask is constructed from the regressed anchor points as a post-processing step during inference. For the matching methods Nearest Point and Nearest Line, we choose an arbitrary point as the origin and connect adjacent points sequentially. For Corner Point with Projection, a mask is similarly constructed from only the valid points. 

\subsection{PointSetNet}

\begin{figure*}[t]
\begin{center}
\includegraphics[width=0.95\linewidth]{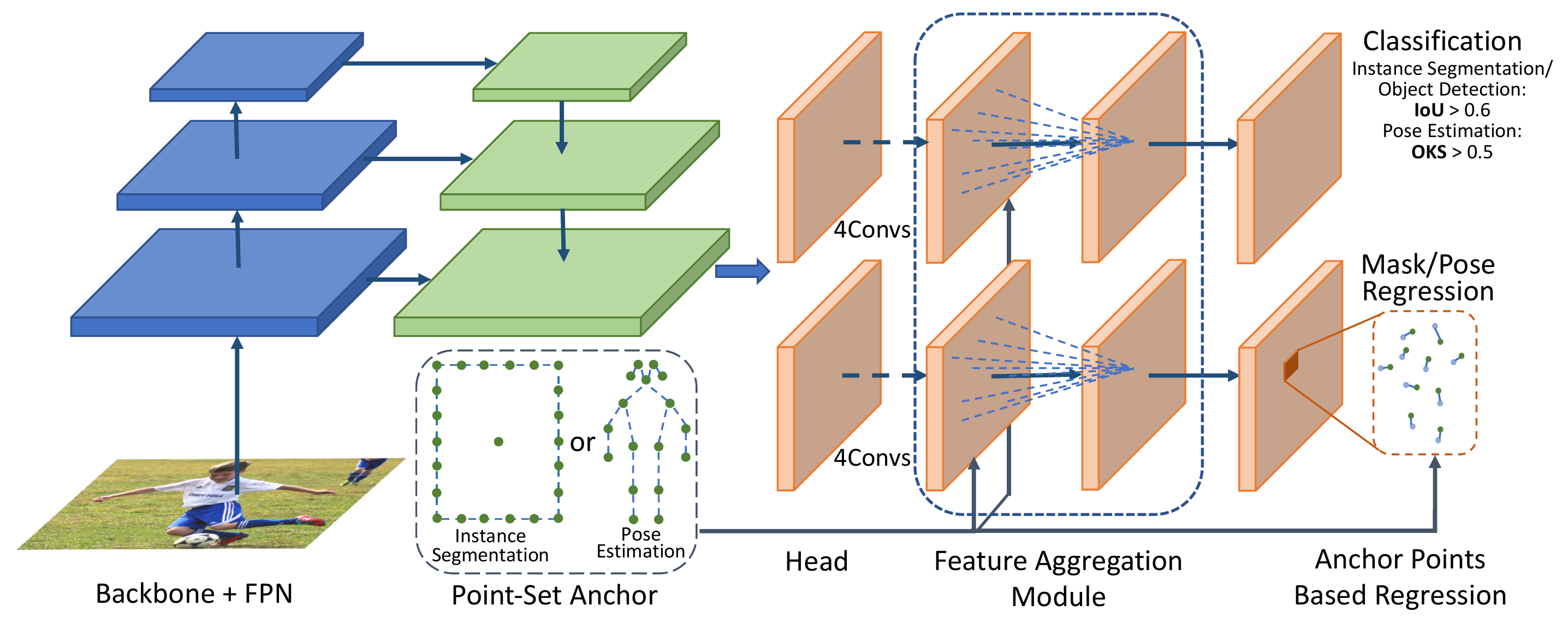}
\end{center}
\vspace{-2ex}
\caption{Network architecture. The left part represents the backbone and feature pyramid network to extract features from different levels. The right part shows the shared heads for classification and mask/pose estimation with point-set anchors. We omit the bounding box regression branch for clearer illustration.}
\label{fig:network}
\vspace{-0.5cm}
\end{figure*}

\noindent\textbf{Architecture.}
PointSetNet is an intuitive and natural extension of RetinaNet~\cite{lin2017focal}. Conceptually, it simply replaces the classical rectangular anchor with the proposed point-set anchor, and attaches a parallel regression branch for instance segmentation or pose estimation in the head. Fig.~\ref{fig:network} illustrates its network structure. Following RetinaNet, we use a multi-scale feature pyramid for detecting objects at different scales. Specifically, we make use of five levels of feature maps, denoted as $\{P_3, P_4, P_5, P_6, P_7\}$. $P_3$, $P_4$ and $P_5$ are generated by backbone feature maps $C_3$, $C_4$ and $C_5$ followed by a $1\times1$ convolutional layer and lateral connections as in FPN~\cite{lin2017feature}. $P_6$ and $P_7$ are generated by a $3\times3$ stride-2 convolutional layer on $C_5$ and $P_6$, respectively. As a result, the stride of $\{P_3, P_4, P_5, P_6, P_7\}$ is $\{8, 16, 32, 64, 128, 256\}$. The head contains several subnetworks for classification, mask or pose regression, and bounding box regression. Each subnetwork contains four $3\times3$ stride-1 convolutional layers, a feature aggregation layer which is used only for the pose estimation task, and an output layer. Table~\ref{tab:out_layers} lists the output dimensions from the three subnetworks for instance segmentation and pose estimation. Following~\cite{lin2017feature,lin2017focal}, we also share the head among $P3-P7$.

\begin{table}[ht]
\center
\caption{Output dimensions of different subnetworks for instance segmentation and pose estimation. $K$, $n$ and $C$ denote the number of point-set anchors, number of points per point-set anchor, and class number for the training/testing dataset, respectively.} 
\begin{tabular}{c|c|c|c}
\hline
Task  & Classification & Shape Regression & Bounding Box Regression\\
\hline
Instance Segmentation & $K\times C$ & $K\times(n\times2)$ & $K\times 4$\\
Pose Estimation & $K\times2$ & $K\times(17\times2)$ & -\\
\hline
\end{tabular}
\label{tab:out_layers}
\end{table}

\noindent\textbf{Point-set anchor density.}
One of the most important design factors in anchor based detection frameworks~\cite{renNIPS15fasterrcnn,lin2017feature,lin2017focal,liu2016ssd} is how densely the space of possible anchors is sampled. For instance segmentation, following~\cite{lin2017focal}, we simply replace classical rectangular anchors by our mask point-set anchors, and use 3 scales and 3 aspect ratios per location on each of the feature maps. Specifically, we make use of the implicit bounding box in point-set anchors, where each bounding box has three octave scales $2^{k/3}(k\leq3)$ and three aspect ratios $[0.5, 1, 2]$, and the base scale for feature maps $P3-P7$ is $\{32, 64, 128, 256, 512\}$. The combination of octave scales, aspect ratios and base scales will generate 9 bounding boxes per location on each of the feature maps. Anchor points are uniformly sampled on the four sides of the generated bounding boxes. For pose estimation, we use 3 mean poses generated by the k-means clustering algorithm. Then we translate them to each position of the feature maps as the point-set anchors. We further use 3 scales and 3 rotations for each anchor, yielding 27 anchors per location. The other feature map settings are the same as in instance segmentation.

\noindent\textbf{Loss function.}
We define our training loss function as follows:
\begin{equation}
    L = \frac{1}{N_{pos}}\sum_{x,y}L_{cls}(p_{x,y}, c^*_{x,y}) + \frac{\lambda}{N_{pos}}\sum_{x,y} \mathds{1}_{\{c^*_{x,y}\}>0} L_{reg}(t_{x,y}, t^*_{x,y}),
\end{equation}
where $L_{cls}$ is the Focal loss in~\cite{lin2017focal} and $L_{reg}$ is the L1 loss for shape regression. $c^*_{x,y}$ and $t^*_{x,y}$ represent classification and regression targets, respectively. $N_{pos}$ denotes the number of positive samples and $\lambda$ is the balance weight, which is set to $0.1$ and $10.0$ for instance segmentation and pose estimation, respectively. $\mathds{1}_{\{c^*_{x,y}\}>0}$ denotes the indicator function, being 1 if $\{c^*_{x,y}\}>0$ and 0 otherwise. The loss is calculated over all locations and all feature maps.

\noindent\textbf{Elements specific to pose estimation.}
Besides target normalization and the embedding of prior knowledge in the anchor shapes, we further show how feature aggregation with point-set anchors achieves a certain feature transformation invariance and how point-set anchors can be extended to multi-stage learning.

\noindent\textbf{-- Deep shape indexed features.}
Learning of shape/transformation invariant features has been a fundamental problem in computer vision~\cite{sun2015cascaded,dollar2010cascaded}, as they provide consistent and robust image information that is independent of geometric configuration. A point-set anchor acts as a shape hypothesis of the object to be localized. Though it reflects a coarse estimate of the target object shape, it still achieves a certain feature invariance to object shape,
as it extracts the feature in accordance with the ordered point-set. The blue dashed rectangle in Fig.~\ref{fig:network} depicts the feature aggregation module. In principle, the closer the anchor points are to the object shape, the better shape invariance of the feature. The deep shape indexed feature is implemented by DCN~\cite{dai2017deformable}. Specifically, we replace the learnable offset in DCN with the location of points in a point-set anchor.

\noindent{\textbf{-- Multi-stage refinement.}}
Holistic shape regression is generally more difficult than part based heat map learning~\cite{sun2018integral}. This is on one hand because of the large and continuous solution space of poses. On the other hand, it is due to the extremely unbalanced transformation variance between different keypoints. To address this, a classic paradigm is to estimate the pose progressively via a sequence of weak regressors where each weak regressor uses features that depend on the estimated pose from the previous stage~\cite{dollar2010cascaded,sun2015cascaded}.  

To this end, we use an additional refinement stage for pose estimation. While the $k$ mean poses in the training set are used as the initial anchors for the first stage, we can directly use the pose predictions of the first stage as the point-set anchors for the second stage. Since the joint positions in the point-set anchors are well-initialized in the first stage, the point-set anchors for the second stage are much closer to the ground truth shapes. This facilitates learning since the distance of the regression target becomes much smaller and better shape-invariant features can be extracted by using the more accurate anchor shapes.



Conceptually, this head network can be stacked for multi-stage refinement, but we find the use of a single refinement to be most effective. Hence, we use one-step refinement for simplicity and efficiency.

\section{Experiments}
\label{sec:exp}

\subsection{Instance Segmentation Settings}

\noindent\textbf{Dataset.} We present experimental results for instance segmentation and object detection on the MS-COCO~\cite{lin2014microsoft} benchmark. We use COCO \texttt{trainval35k} (115k images) for training and the \texttt{minival} split (5k images) for ablations. Comparisons to the state-of-the-art are reported on the \texttt{test-dev} split (20k images).

\begin{table}[t]
\center
\caption{Comparison of different matching strategies between anchor points and mask contours on the instance segmentation task.} 
\begin{tabular}{C{4.5cm}|C{0.8cm}|C{0.8cm}|C{0.8cm}|C{0.8cm}|C{0.8cm}|C{0.8cm}}
\hline
Matching strategy  & $AP$ & $AP_{50}$ & $AP_{75}$ & $AP_{S}$&$AP_{M}$ &$AP_{L}$\\
\hline
Nearest Point & 21.9 & 42.1& 20.6& 11.4& 24.6& 29.8\\
Nearest Line & 23.2& 46.5& 21.0& 12.5& 26.2&32.0\\
Corner Point with Projection & \textbf{27.0}&\textbf{49.1} &\textbf{26.6} &\textbf{13.8} &\textbf{30.6} &\textbf{36.7}\\
\hline
\end{tabular}
\label{tab:segm_match}
\end{table}

\noindent\textbf{Training details.} All our ablation experiments, except when specifically noted, are conducted on the \texttt{minival} split with ResNet-50~\cite{he2016deep} and FPN. Our network is trained with synchronized stochastic gradient descent (SGD) over 4 GPUs with a mini-batch of 16 images (4 images per GPU). We adopt the $1\times$ training setting, with 12 epochs in total and the learning rate initialized to 0.01 and then divided by 10 at epochs 8 and 11. The weight decay and momentum parameters are set to $10^{-4}$ and $0.9$, respectively. We initialize our backbone network with the weights pre-trained on ImageNet~\cite{deng2009imagenet}. For the newly added layers, we keep the same initialization as in~\cite{lin2017focal}. Unless specified, the input images are resized to have a shorter side of 800 and a longer side less than or equal to 1333. Samples with IoU higher than 0.6 and lower than 0.4 are defined as positive samples and negative samples, respectively. We use the Corner Point with Projection matching strategy and set the number of anchor points to 36 by default. 

\noindent\textbf{Inference details.}
We forward the input image through the network and obtain the classification scores and the corresponding predicted classes. According to the classification scores, the top-1k anchors from each level of the feature maps are sent for mask construction. Then the top predictions from all levels are merged and non-maximum suppression (NMS)\footnote{IoU is calculated from predicted bounding boxes and ground-truth rectangles due to the high computation of mask IoU. If there is no bounding box branch in the network, we use the smallest rectangle that encompasses the predicted mask instead.} with a threshold of 0.5 is employed as post-processing.

\subsection{Experiments on Instance Segmentation}

\begin{table}[t]
\center
\caption{Comparison on different numbers of anchor points.} 
\begin{tabular}{C{1cm}|C{0.9cm}|C{0.9cm}|C{0.9cm}|C{0.9cm}|C{0.9cm}|C{0.9cm}}
\hline
n& $AP$ & $AP_{50}$ & $AP_{75}$& $AP_{S}$&$AP_{M}$ &$AP_{L}$\\
\hline

24 & 26.7&	48.8&	26.1&	13.6&	30.2&	36.4\\
36 & 27.0&	49.1&	26.6&	13.8&	30.6&	36.7\\
48 & 27.2 & 49.2&	26.8&	13.9&	30.7&	36.7\\
60 & 28.0 & \textbf{49.8}&	27.9&	13.9&	31.4&	38.6\\
72 & \textbf{28.0} & 49.6&	\textbf{27.9}&	\textbf{14.6}&	\textbf{31.5}&	\textbf{38.8}\\
\hline
\end{tabular}
\label{tab:segm_TP_num}
\end{table}

\noindent\textbf{Mask matching strategies.}
First, we compare the results from the three matching strategies between a point-set anchor and the corresponding mask contour as shown in Table~\ref{tab:segm_match}. Nearest Point has the worst performance, perhaps because each polygon point may be assigned to multiple anchor points and this inconsistency may misguide training. Both Nearest Line and Corner Point with Projection treat the ground-truth mask as a whole contour instead of as discrete polygon points in the mask matching step. However, there still exist ambiguous anchor points for the Nearest Line method as shown in Fig.~\ref{fig:match_methods}(b). Corner Point with Projection eliminates inconsistency, as the subdivision of the mask contour into segments leads to better-defined matches, and it achieves the best performance among the three methods.

\begin{table}[t]
\center
\caption{Comparison of two regression origins.} 
\begin{tabular}{c|C{0.9cm}|C{0.9cm}|C{0.9cm}|C{0.9cm}|C{0.9cm}|C{0.9cm}}
\hline
Origin  & $AP$ & $AP_{50}$ & $AP_{75}$& $AP_{S}$&$AP_{M}$ &$AP_{L}$\\
\hline
Center Point & 26.0 & 48.4& 24.8 &13.6  &29.3 &35.4\\
Anchor Points &  \textbf{27.0}&	\textbf{49.1}&	\textbf{26.6}&	\textbf{13.8}&	\textbf{30.6}&	\textbf{36.7}\\
\hline
\end{tabular}
\label{tab:reg_methods}
\end{table}

\noindent\textbf{Effect of point-set anchors.}
The number of anchor points can greatly affect instance segmentation. From Table~\ref{tab:segm_TP_num}, it can be seen that the more accurate representation from more anchor points leads to better performance. Also, the performance is seen to saturate beyond a certain number of anchor points, e.g., 72 points. We also demonstrate the benefits of using point-set anchors as the regression origin as shown in Table~\ref{tab:reg_methods}. It can be seen that with the same regression targets, using point-set anchors as the regression origin outperforms center point based regression by 1.0 AP.

\noindent\textbf{Comparison with state-of-the-art methods.}
We evaluate PointSetNet on the COCO \texttt{test-dev} split and compare with other state-of-the-art object detection and instance segmentation methods. We use 60 anchor points for contour representation and ResNext-101 with DCN~\cite{dai2017deformable} as backbone. For data augmentation, we randomly scale the shorter side of images in the range of 480 to 960 during training and increase the number of training epochs to 24 ($2\times$ training setting). An image pyramid with a shorter side of $\{400, 600, 800, 100, 1200\}$ is applied during inference. As shown in Table~\ref{tab:segm_comparison}, 
PointSetNet achieves performance competitive with the state-of-the-art in both object detection and instance segmentation. The gap between TensorMask and PointSetNet arises from the tensor bipyramid head which brings a +5.1 AP improvement. We do not plug this into our framework due to its heavy time and memory cost. Although the AP is 0.2 lower than PolarMask on instance segmentation, our proposed point-set anchor has the benefit of being applicable to human pose estimation.

\begin{table}[t]
\scriptsize
\center
\renewcommand\tabcolsep{5.0pt}
\caption{Results on the MS COCO \texttt{test-dev} compared to state-of-the-art \textbf{instance segmentation} and \textbf{object detection} methods. `$*$' denotes multi-scale testing.} 
\begin{tabular}{l|l|C{1.2cm}|C{0.5cm}|C{0.5cm}|C{0.5cm}|C{0.5cm}|C{0.5cm}|C{0.5cm}}
\hline
\multirow{2}{*}{Method} & \multirow{2}{*}{Backbone} &
Regression & \multicolumn{3}{c|}{Segmentation} & \multicolumn{3}{c}{Detection} \\
\cline{4-9}
 ~& ~& Based &${AP}$ & ${AP_{50}}$ & ${AP_{75}}$ & ${AP}$ & ${AP_{50}}$ & ${AP_{75}}$\\
\hline
\hline
Mask RCNN~\cite{he2017mask} & ResNeXt-101 &\xmark &37.1 & 60.0& 39.4& 39.8& 62.3&43.4 \\
\hline
TensorMask~\cite{chen2019tensormask} & ResNet-101& \xmark&37.1 &59.3 & 39.4&- &- &-  \\
\hline
FCIS~\cite{li2017fully} & ResNet-101 &\xmark & 29.2 & 49.5& -& -& -& - \\
\hline
YOLACT~\cite{bolya2019yolact} & ResNet-101& \xmark&31.2 & 50.6&32.8 &33.7 & 54.3& 35.9 \\
\hline
ExtremeNet$^*$\cite{zhou2019bottom} & Hourglass-104&\xmark & 18.9& 44.5& 13.7&43.7 &60.5 & 47.0 \\
\hline
CornerNet$^*$~\cite{law2018cornernet} & Hourglass-104 & \xmark & - & - & - & 42.2 & 57.8& 45.2 \\
\hline
RetinaNet~\cite{lin2017focal} & ResNext-101 &\checkmark & -&- &-&40.8 & 61.1& 44.1 \\
\hline
FCOS~\cite{tian2019fcos} & ResNext-101&\checkmark &- & -&- &44.7 & 64.1& 48.4 \\
\hline
CenterNet$^*$\cite{zhou2019objects} & Hourglass-104 &\checkmark & -& -& -& 45.1& 63.9& 49.3 \\
\hline
RepPoints$^*$\cite{yang2019reppoints} & ResNeXt-101-DCN &\checkmark &- &- &- &46.5 &67.4 & 50.9 \\
\hline
PolarMask~\cite{xie2019polarmask} & ResNeXt-101-DCN & \checkmark&36.2& 59.4& 37.7& -& -& - \\
\hline
Dense RepPoints~\cite{yang2019dense} & ResNeXt-101-DCN & \checkmark& 41.8& 65.7& 45.0& 48.9& 69.2&  53.4\\
\hline
ATSS$^*$~\cite{zhang2020bridging} & ResNeXt-101-DCN & \checkmark& -& - & - & 50.7 & 68.9 &  56.3\\
\hline
\hline
Ours & ResNeXt-101-DCN & \checkmark&34.6 & 60.1& 34.9 & 45.1& 66.1& 48.9 \\

Ours$^*$ & ResNeXt-101-DCN & \checkmark&36.0 & 61.5& 36.6 & 46.2& 67.0& 50.5 \\
\hline
\end{tabular}
\label{tab:segm_comparison}
\end{table}

\subsection{Pose Estimation Settings}
\noindent\textbf{Dataset.} We conduct comprehensive comparison experiments on the MS COCO Keypoint dataset~\cite{lin2014microsoft} to evaluate the effectiveness of point-set anchors on multi-person human pose estimation. The COCO train, validation, and test sets contain more than 200k images and 250k person instances labeled with keypoints. 150k of the instances are publicly available for training and validation. Our models are trained only on the COCO \texttt{train2017} dataset (which includes 57K images and 150K person instances) with no extra data. Ablations are conducted on the \texttt{val2017} set, and final results are reported on the \texttt{test-dev2017} set for a fair comparison to published state-of-the-art results~\cite{cao2017realtime,fang2017rmpe,he2017mask,newell2017associative,papandreou2018personlab,papandreou2017towards,zhou2019objects,nie2019single}.

\noindent\textbf{Training details.} For pose estimation, we use the Adam~\cite{kingma2014adam} optimizer for training. The base learning rate is 1e-4. It drops to 1e-5 at 80 epochs and 1e-6 at 90. There are 100 epochs in total. Samples with OKS higher than 0.5 and lower than 0.4 are defined as positive samples and negative samples, respectively. The other training details are the same as for instance segmentation including the backbone network, network initialization, batch size and image resolution.

\subsection{Experiments on Pose Estimation}

\begin{table}[t]
\center
\caption{Comparing different point-set anchors.
Results are better when anchors more \emph{efficiently} cover ground truth shapes with larger positive/negative sample ratio.} 
\begin{tabular}{C{3.4cm}|C{2.5cm}|C{2.0cm}|C{1.1cm}|C{1.1cm}|C{1.1cm}}
\hline
Anchor Type& Matched GT(\%) & Pos/Neg(\permille) & $AP$ & $AP_{50}$ & $AP_{75}$\\
\hline
Center Point &  0 & $4.5$ &  16.9&48.7&6.0\\
Rectangle & 4.0 & $7.7$ & 12.0 & 33.9 & 5.9 \\
Mean Pose & 18.0 & $6.6$& \textbf{40.9} & \textbf{69.4} & \textbf{42.0} \\
\hline
K-means\_3 & 27.7 &$6.4$&43.3&70.7&45.5 \\
K-means\_5 & 32.1 &$6.1$& \textbf{43.8} &\textbf{71.3}&\textbf{46.6} \\
K-means\_7 & 34.6 &$6.0$& 43.8 & 71.2 & 46.5\\
\hline
Scale (0.8:0.2:1.2) & 25.2 & $6.5$ &   42.6&69.8& 44.9 \\
Scale (0.6:0.2:1.4) & 27.6 & $6.2$ & \textbf{42.6} & \textbf{69.7} & \textbf{44.5}\\
Scale (0.4:0.2:1.6) & 29.6 & $5.9$ &   39.4&65.4&41.2 \\
\hline
Rotation (-10:10:10) &27.2 & $9.0$ & 42.6 & 70.6 & 44.2\\
Rotation (-20:10:20) &36.0  & $6.6$ & \textbf{43.5}&\textbf{71.6}&\textbf{45.9}\\
Rotation (-30:10:30) &40.5  & $6.2$  & 42.6&70.6&44.4 \\
\hline
\end{tabular}
\label{tab:pose_template_anchor}
\end{table}

\noindent\textbf{Effect of point-set anchors.} We first compare the proposed point-set anchors with strong prior knowledge of pose shapes to other point-based anchors like the center point and points on a rectangle. We denote as \emph{mean pose} the use of the average pose in the training set as the canonical shape. Then we translate this \emph{mean pose} to every position in the image as the point-set anchors for pose estimation. No other transformation is used to augment the anchor distribution for a fair comparison to the \emph{center point} and \emph{rectangle} anchors. 
A ground truth pose is assigned to anchors with OKS higher than 0.5. If no anchor is found higher than OKS 0.5, the ground truth is assigned to the closest anchor.

In Table~\ref{tab:pose_template_anchor}, it is shown that the mean pose anchor outperforms the \emph{center point} and \emph{rectangle} anchors by a large margin with more ground truth poses assigned with OKS greater than 0.5. Specifically, it surpasses \emph{center point} anchors by 24 AP and \emph{rectangle} anchors by 28.9 AP. This indicates that an anchor that better approximates the target shape is more effective for shape regression.

\begin{table}[t]
\center
\caption{Comparing the deep shape indexed feature with other feature extraction methods. Deep-SIF-$n$ denotes deep shape indexed feature with $n$ points used for feature extraction.} 
\begin{tabular}{C{4.3cm}|C{1.3cm}|C{1.3cm}|C{1.3cm}|C{1.3cm}|C{1.3cm}}
\hline
Feature Types& Loss\_cls & Loss\_reg & $AP$ & $AP_{50}$ & $AP_{75}$\\
\hline

Center Feature &0.31 & 5.92& 40.9 & 69.4 & 42.0\\
Box Corner Feature &0.31&6.22&42.6&68.2&45.6\\
Box Region Feature &0.30&5.99& 42.8 & 69.0 & 46.0\\ 
\hline
Deep-SIF-9 for Cls &0.30&6.29&41.6&68.2&44.0\\
Deep-SIF-9 for Reg &0.32&5.93&45.5&69.6&49.1\\
Deep-SIF-9 for Cls $\&$ Reg &0.29&5.92&46.0&71.7&49.1\\\
Deep-SIF-25 for Cls $\&$ Reg &\textbf{0.29}&\textbf{5.79}&\textbf{47.5}&\textbf{72.0}&\textbf{51.6}\\
\hline
\end{tabular}
\label{tab:pose_shape_indexed_feature}
\end{table}

We obtain further improvements by using additional canonical pose shapes generated by the K-Means clustering algorithm or by augmenting the \emph{mean pose} shape with additional sampling of rotation and scaling transformations. However, more anchor shapes and transformation augmentations also introduce more negative anchors that are not assigned to any of the ground truth poses, which makes learning less efficient. Better performance can be attained with a better trade-off between covering more ground truth shapes and incurring fewer negative samples. Empirically, we achieve the best performance by using 5 canonical pose shapes (+2.9AP), 5 scale transformations (+1.7AP) and 5 rotation transformations (+2.6AP).




\noindent\textbf{Effect of deep shape indexed feature.} Table~\ref{tab:pose_shape_indexed_feature} compares the deep shape indexed feature with other feature extraction methods. First, three regular feature extraction methods, namely from the center point, 4 corner points of the bounding box and 9 grid points in the bounding box region are compared. With more feature extraction points, slightly better performance is obtained. 

Then, we use the deep shape indexed feature which uses the point set in the anchors for feature extraction. A point set based on pose shape priors extracts more informative features that greatly improve learning and performance. Specifically, with the same number of points (part of the 17 joints) as in the box region feature (i.e., 9), the deep shape indexed feature improves the AP from 42.8 to 46.0 (+3.2 AP, relative 7.5\% improvement). Further improvement can be obtained by using more joint points for feature extraction. Note that if the shape indexed feature is used only for the person classification sub-network, the improvement is not as great as using it only in the pose regression sub-network. This indicates that it mainly enhances pose regression learning.


\begin{table}[t]
\center
\caption{Effect of multi-stage refinement.} 
\begin{tabular}{c|c|c|c|c|c|C{0.8cm}|C{0.8cm}|C{0.8cm}}
\hline
Stage & OKS &Matched GT(\%)&Pos/Neg(\permille)&  Loss\_cls & Loss\_reg & $AP$ & $AP_{50}$ & $AP_{75}$\\
\hline
Stage-1 &0.5&45.7&12.7&0.25&5.52& 48.3 & 74.3 & 52.5\\
\hline
Stage-2 &0.99&81.5&11.5&0.23 & 4.28& 58.0 & 80.8 & 62.4 \\
\hline
\end{tabular}
\label{tab:pose_multi_stage}
\end{table}

\begin{table}[t]
\center
\caption{Effect of backbone network and multi-scale testing.} 
\begin{tabular}{C{3.2cm}|C{1.7cm}|C{1.0cm}|C{1.0cm}|C{1.0cm}}
\hline
Backbone & Multi-Test & $AP$ & $AP_{50}$ & $AP_{75}$\\
\hline
ResNet-50 &\xmark&      58.0 & 80.8 & 62.4          \\
\hline
ResNeXt-101-DCN & \xmark & 62.5&83.1&68.3          \\
\hline
ResNeXt-101-DCN &\checkmark&  65.7    &   85.4    &   71.8  \\
\hline
HRNet-W48 &\checkmark&  69.8    &  88.8    &   76.3  \\
\hline
\end{tabular}
\label{tab:backbone_multitest}
\end{table}

\noindent\textbf{Effect of multi-stage refinement.} Table~\ref{tab:pose_multi_stage} shows the result of using a second stage for refinement. Since the anchors for the second stage are much closer to the ground truth, we use a much higher OKS threshold (0.99) for positive sample selection. Even though the second stage uses a much higher OKS threshold, it is found that many more ground truth poses are covered. Both the person classification and the pose regression losses are decreased. We thus obtain significant improvement from the second stage, specifically, 9.7 AP (relative 20.1\% improvement) over the first stage.



\noindent\textbf{Effect of stronger backbone network and multi-scale testing.} Table~\ref{tab:backbone_multitest} shows the result of using stronger backbones and multi-scale testing. Specifically, we obtain 4.5 AP improvement from using the ResNeXt-101-DCN backbone network. A 3.2 AP improvement is found from using multi-scale testing. We obtain further improvement by using HRNet~\cite{sun2019deep} as the backbone (+4.1 AP). 

\noindent\textbf{Comparison with state-of-the-art methods.} Finally, we test our model (with HRNet backbone and multi-scale testing) on the MSCOCO \texttt{test-dev2017} dataset and compare the result to other state-of-the-art methods in Table~\ref{table:sota}. PointSetNet outperforms CenterNet~\cite{zhou2019objects} by 5.7 AP and achieves competitive results to the state-of-the-art.

\begin{table}[t]
\caption{Results on the MS COCO \texttt{test-dev2017} compared to state-of-the-art \textbf{pose estimation} methods.}
\begin{center}
\begin{tabular}{ L{2.8cm} |L{3.8cm} | C{0.9cm}  C{0.9cm}  C{0.9cm}  C{0.9cm}  C{0.9cm}  }
\hline
Method & Backbone  & $AP$ & $AP_{50}$ & $AP_{75}$ & $AP_{M}$ & $AP_{L}$ \\
\hline
\hline
\emph{Heat Map Based}&  &  &   &  &  &   \\
CMU-Pose~\cite{cao2017realtime}& 3CM-3PAF (102) & 61.8 &  84.9 & 67.5 & 57.1 &  68.2 \\
RMPE ~\cite{fang2017rmpe}&Hourglass-4 stacked & 61.8  &83.7 & 69.8  & 58.6 &67.6 \\
Mask-RCNN ~\cite{he2017mask}& ResNet-50 & 63.1 & 87.3 & 68.7 & 57.8 & 71.4\\
G-RMI ~\cite{papandreou2017towards}&ResNet-101+ResNet-50  &  64.9   & 85.5 & 71.3 & 62.3 & 70.0\\
AE~\cite{newell2017associative}&Hourglass-4 stacked  & 65.5    & 86.8 & 72.3 & 60.6 & 72.6\\
PersonLab~\cite{papandreou2018personlab}& ResNet-152 &  68.7   &89.0  & 75.4 &64.1  & 75.5\\
HigherHRNet~\cite{cheng2019bottom}&HRNet-W48&70.5 &89.3 &77.2 &66.6 &75.8\\
CPN~\cite{chen2018cascaded} & ResNet-Inception & 72.1& 91.4& 80.0& 68.7& 77.2\\
SimpleBaseline~\cite{xiao2018simple}&  ResNet-152 &73.7 &91.9 &81.1 &70.3 &80.0\\
HRNet~\cite{sun2019deep,wang2020deep}&HRNet-W48&75.5 &92.5 &83.3 &71.9 &81.5\\
DARK~\cite{zhang2020distribution}&HRNet-W48&76.2 &92.5& 83.6& 72.5& 82.4\\
\hline
\emph{Regression Based}&  &  &   &  &  &   \\
CenterNet~\cite{zhou2019objects}& Hourglass-2 stacked (104) &   63.0  &86.8  &69.6  &58.9  & 70.4\\
SPM~\cite{nie2019single}&Hourglass-8 stacked &  66.9   & 88.5  & 72.9  & 62.6 & 73.1 \\
Integral \cite{sun2018integral}& ResNet-101  & 67.8 & 88.2 & 74.8    & 63.9 & 74.0   \\
Ours & HRNet-W48 &  68.7  & 89.9 & 76.3 & 64.8  & 75.3\\
\hline
\end{tabular}
\vspace{-1cm}
\end{center}
\label{table:sota}
\end{table}
\vspace{-0.4cm}
\section{Conclusion}
\vspace{-0.4cm}
In this paper, we propose Point-Set Anchors which can be seen as a generalization and extension of classical anchors for high-level recognition tasks such as instance segmentation and pose estimation. Point-set anchors provide informative features and good task-specific initializations which are beneficial for keypoint regression. Moreover, we propose PointSetNet by simply replacing the anchor boxes with the proposed point-set anchors in RetinaNet and attaching a parallel branch for keypoint regression. Competitive experimental results on object detection, instance segmentation and human pose estimation show the generality of our point-set anchors.




%
%
\bibliographystyle{splncs04}
\bibliography{egbib}
\end{document}